# Unsupervised Cross-Media Hashing with Structure Preservation

Xiangyu Wang
Institute for Infocomm Research

Alex Yong-Sang Chia
Institute for Infocomm Research

## Abstract

*Recent years have seen the exponential growth of heterogeneous multimedia data. The need for effective and accurate data retrieval from heterogeneous data sources has attracted much research interest in cross-media retrieval. Here, given a query of any media type, cross-media retrieval seeks to find relevant results of different media types from heterogeneous data sources. To facilitate large-scale cross-media retrieval, we propose a novel unsupervised cross-media hashing method. Our method incorporates local affinity and distance repulsion constraints into a matrix factorization framework. Correspondingly, the proposed method learns hash functions that generates unified hash codes from different media types, while ensuring intrinsic geometric structure of the data distribution is preserved. These hash codes empower the similarity between data of different media types to be evaluated directly. Experimental results on two large-scale multimedia datasets demonstrate the effectiveness of the proposed method, where we outperform the state-of-the-art methods.*

## 1. Introduction

The proliferation and pervasiveness of social networks have contributed much to the exponential growth in the amount of user-generated multimedia content. For example, there are about 500 million tweets sent each day, of which 36% of the tweets contain images [1]. Given that information is often presented in different media types (e.g. text and images), it is desirable to have a cross-media retrieval paradigm to search large scale heterogeneous data given queries in any media types. For example, such a retrieval paradigm would find relevant textual content when provided with an image query, and vice versa. Here, other than the issue of retrieval accuracy, two important considerations for practical usage are the storage and computational requirements of the retrieval system.

Hashing-based nearest neighbor search has received considerable interest for their great efficiency gains in massive data [9]. A recent trend in cross-media retrieval is based on the concept of locality-sensitive hashing (LSH) [10]. Specifically, given data of different modalities, such methods learn hash functions from the data automatically. These hash functions are then exploited to transform the data points (of different modalities) from different feature space into a common Hamming space. Retrieval of similar data that are of different modalities can then be achieved in this common space via nearest neighbor search. Such hash functions are often engineered to store the hash codes in binary bits, and linear projection hash functions are usually preferred because of their efficiency in hashing time. This further improves the query speed, and affords more effective storage requirements.

Recently, there have been a few works on cross-media hashing that have achieved notable success [17, 23, 8, 7]. Zhen et al. [23] and Gong et al. [8] exploited the relationships between text-image pairs to compute the hash functions. These methods consider only the inter-media structure of the data, but ignore the intra-media structure [2] (i.e. relationships between image-image and text-text pairs). Ding et al. [7] learned unified hash codes for corresponding instances of different modalities (i.e. image-text pairs). Like [23, 8], their method minimizes inter-media loss, but does not explicitly consider intra-media structure of the data. To exploit more data information, Song et al. [17] exploited both inter-media and intra-media data consistency to optimize the hash functions. While good results were reported by these previous works, we note that they optimize the hash functions based on local geometrical properties of the data but neglect the global geometrical data structure, which needs to be respected in many applications [18].

Hashing methods can be divided into two different types based on the extent of supervision required during the learning phase. Specifically, supervised methods such as Co-Regularized Hashing [23] and Semantic Correlation Maximization Hashing [22] require prior knowledge such as semantic labels of the features (i.e. class labels) to learn the hash functions. Given that most multimedia data have lim-ited or no semantic labels, such methods demand expensive and laborious work to obtain clean annotated data. This limits its use in practical applications. Unsupervised hashing methods, on the other hand, do not require such semantic labels and only require correspondences between pairs of data (i.e. knowledge of image-text pairs) to be known during the learning phase. Such information is more readily available (e.g., an image tweet with accompanying hashtags/captions), and hence unsupervised methods are more popular and suitable for real-world applications.

---

[1] http://socialtimes.com/is-the-status-update-dead-36-of-tweets-are-photos-infographic_b103245

[2] In this paper, we use the terms "data structure" and "data manifold" interchangeably.

In this work, we present a novel unsupervised cross-media hashing method which is motivated on the following key concepts. First, data points of different modalities but with similar semantics should be projected near to each other in the common Hamming space, i.e., hash functions should be trained to minimize inter-media loss. This ensures more reliable cross-media retrieval ability with nearest neighbor search. Second, data points which are similar in the original feature space should also be similar in the projected space. In the same aspect, data points which are far apart in the original feature space should also be projected to be far apart in the projected space. This retains the intra-media structure of different modalities after projecting into a common Hamming space. By preserving such distances between data points (intra-media structure), information and discriminability of the original data are also retained. We are aware of some works [17] that exploit nearest neighbor cues, which ensure similar data to be close together, to preserve the structure of the intra-media space. However, given that they neglect the structure of dissimilar data in the optimization, data points which are dissimilar might be projected close to each other in the common space and be erroneously retrieved with nearest neighbor search.

We illustrate these key concepts in Fig. 1. Specifically, the top left and bottom left axes represent the image space and text space respectively, and the right axes denote the common Hamming space. Dashed colored lines depict images and text with similar semantics. We formulate the cross-media hashing method as a loss function minimization problem. Here, we formulate the loss function to incorporate local affinity and distant repulsion constraints in a matrix factorization framework. This maintains the inter-media correspondence and preserves intra-media structure of the data, as shown in the right axes in Fig. 1. To our knowledge, this is the first attempt to explicitly preserve the structure of the data in unsupervised cross-media hashing. Experimental results on two large-scale datasets demonstrate the effectiveness of our method, where we outperform the state-of-the-art methods on these datasets.

## 2. Related work

Hashing methods seek to learn binary representation of data and have been studied extensively. These methods can be categorized based on the modality of data used as single-media hashing and cross-media hashing. Single-media hashing methods focus on the generation of hash functions for data belonging to the same media type. Spectral hashing [20] is one of the most popular hashing methods. It seeks compact binary codes so that the Hamming distance between hashing codes correlates with feature similarity. Wang et al. [19] presented a sequential projection hashing method which learns the hash functions by maximizing the empirical accuracy with an information-theoretic regularizer. Liu et al. [14] proposed a kernel-based supervised hashing method which sequentially trains the hash functions based on similar and dissimilar pairs. Following that, an iterative quantization algorithm [8] is used to transform the original data to the vertices of a zero-centered binary hypercube by minimizing the quantization error of the transformation.

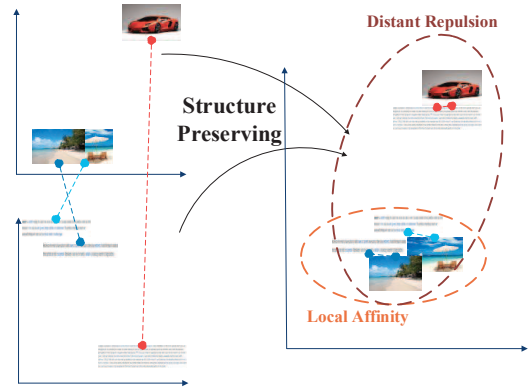

Figure 1: Illustration of the key concepts in the proposed method. Top left axes denote the image space, with example images, while bottom left axes depict the text space and example text articles. Right axes depict the common Hamming space. Dashed colored lines depict images and text with similar semantics. By enforcing distant repulsion and local affinity constraints, the proposed method aims to preserve the structure of the data.

Recently, there have been a few works on cross-media hashing, such as [17, 23, 12, 2, 7], in which the data used are of different modality types. These methods can be divided into supervised and unsupervised hashing methods. Supervised methods require semantic labels of the features. For example, Zhen et al. [23] modeled inter-modality loss by a smoothly clipped inverted squared deviation function, and introduced a hinge loss function so that the hash functions have good generalization ability. A boosting procedure is then used to minimize the bias introduced by the hash functions. Gong et al. [8] developed a cross-media retrieval method which maximizes the correlation between the corresponding data points in different modalities based on canonical correlation analysis. By regarding label vectors as the third modality, their method incorporates semantic labels into the canonical correlation analysis. Zhang et al. [22] proposed a supervised multimodal hashing method where they maximized the semantic correlation among different modalities. Similar to [23, 8], their method demands semantic labels of the training data. Bronstein et al. [2] proposed a cross-modality similarity-sensitive hashing where they learned two groups of linear hash functions to preserve semantic similarity among heterogeneous data pairs. AdaBoost procedure is then used to generate multiple bits sequentially. While good results are obtained by [2, 23, 8, 22], we note that these methods demand semantic (i.e. class) labels of the training data, which may not be readily available. This limits its applicability.

Unsupervised cross-media hashing methods do not require semantic labels during training. Given that unlabelled data are more commonly available, such unsupervised hashing methods are often more suitable for practical applications. Song *et al.* [17] proposed inter-media hashing which utilized the consistency within each media type to learn the hash functions. Their algorithm maintains inter-media consistency and intra-media consistency, where similar data in different and same modalities are projected close to each other in Hamming space. Yu et al. [21] extended this with a kernelized hashing method. While improved results were obtained, their method is less efficient as the query cannot be encoded off-line. In [7], Ding et al. learned unified latent representation for corresponding data in different modalities through collective matrix factorization. Their latent representations preserve the similarity with a loose bound that can be significantly affected by reconstruction errors. Consequently, the intra-media structure is not well preserved by their method.

We note that current generation of cross-media hashing algorithms exploit local geometries of the data to learn the hash functions. Specifically, such functions are designed to ensure semantically related data points are projected close together in the Hamming space. However, these methods ignore the extent of dissimilarity between data points, and hence global data structure is not preserved in the optimization. Consequently, data points which are dissimilar may be projected close together, and hence be erroneously retrieved. In this paper, we learn linear hash functions by explicitly preserving inter-media and intra-media structure of the data and show that this improves the cross-media retrieval performance.

## 3. Structure Preserving Collective Matrix Factorization Hashing

This section presents our method which exploits the correspondence information between data of different modalities to learn the optimal hash functions. Such information can be extracted from data pairings e.g. photo and its caption (which are more readily available as compared to semantic, i.e. class, labels of the data), and reveals semantic relationships between data pairings. Here, we require corresponding data of different modalities to have the same latent representation, and leverage on local affinity constraint and distant repulsion constraint to jointly learn the optimal hash functions. These constraints are designed to preserve the global structure of the data (detailed in Sect. 3.2). In this aspect, the proposed method incorporates global structure preservation and informative latent representation into a single framework.

### 3.1. Notations

We first present the notations that would be used in this paper. Let $\mathbf{x}_i$ denote a column vector extracted from $i^{th}$ data instance of modality type $\mathcal{X}$ (e.g. text), where $\mathbf{x}_i \in \mathbb{R}^{D_x}$ and $D_x$ is the number of dimensions of $\mathbf{x}_i$. We define $X = [\mathbf{x}_1, \ldots, \mathbf{x}_N]$ as the data matrix extracted from $N$ data instances of the same modality. Similarly, we define $Y = [\mathbf{y}_1, \ldots, \mathbf{y}_N]$ as the data matrix of $N$ data instances, each of dimension $D_y$, that are extracted from data of a different modality type $\mathcal{Y}$ (e.g. image). Here, $X \in \mathbb{R}^{D_x \times N}$, $Y \in \mathbb{R}^{D_y \times N}$, and $\mathbf{x}_i$ is paired with $\mathbf{y}_i$. Without loss of generality, we assume $\mathbf{x}_i \in X$ and $\mathbf{y}_j \in Y$ are normalized and centered.

For each bit of the hash codes, the hash function for modality $\mathcal{X}$ is defined as $h_x(\mathbf{x}) = sgn(\mathbf{w}_x^T \mathbf{x} + b_x)$, where $sgn(\cdot)$ denotes the sign function, $\mathbf{w}_x \in \mathbb{R}^{D_x}$ is projection vector, and $b_x$ is defined as the mean of the projection. Since the data is centered, $b_x = 0$. Similarly, the hash function for modality $\mathcal{Y}$ is defined as $h_y(\mathbf{y}) = sgn(\mathbf{w}_y^T \mathbf{y} + b_y)$. Here, we seek to learn the set of projection matrices $P_x = [\mathbf{w}_x^{(1)}, \ldots, \mathbf{w}_x^{(H)}]^T$, $P_y = [\mathbf{w}_y^{(1)}, \ldots, \mathbf{w}_y^{(H)}]^T$ which preserve the intra-media structure and inter-media correspondence in the projected space, where $H$ denotes the number of bits in a hash code. For an unseen data instance $\mathbf{x}_p$ of modality $\mathcal{X}$ and $\mathbf{y}_p$ of modality $\mathcal{Y}$, their $t^{th}$ hash bit can be trivially computed as $sgn(\mathbf{w}_x^{(t)^T} \mathbf{x}_p)$ and $sgn(\mathbf{w}_y^{(t)^T} \mathbf{y}_p)$ respectively. With a slight abuse of notation for $sgn(\cdot)$, the hash codes of $\mathbf{x}_p$ and $\mathbf{y}_p$ can then be computed as $sgn(P_x \mathbf{x}_p)$ and $sgn(P_y \mathbf{y}_p)$ respectively.

### 3.2. Structure Preserving Formulation

Inspired by the success of [7] which assigns the same latent representation to data pairings and learns hash codes from such representations, we formulate our method to learn the set of projection matrices $P_x$ and $P_y$ as follows. Let $\mathbf{v}_i$ denote the latent representation of both $\mathbf{x}_i$ and $\mathbf{y}_i$. Correspondingly, $sgn(\mathbf{v}_i)$ is the hash codes of $\mathbf{x}_i$ and $\mathbf{y}_i$. We denote $V = [\mathbf{v}_1, \ldots, \mathbf{v}_N]$ as the latent representations of $X$ and $Y$. The latent representation of $X$ can then be obtained through matrix factorization by minimizing the function $\|X - U_x V\|_F^2$ [6]. Similarly, the latent representation of $Y$ can be obtained by minimizing the function $\|Y - U_y V\|_F^2$. We highlight here that since $V$ is the latent representation of both $X$ and $Y$, we implicitly ensure that data-pairing of different modalities would be mapped onto the same point in the projected space, i.e., their hash codes would be the same. Consequently, this ensures structure of inter-media data is well preserved. Given $P_x$ as the projection matrix of $X$ and $P_y$ as the projection matrix of $Y$, we can derive $P_x$ by minimizing $\|V - P_x X\|_F^2$ and $P_y$ by minimizing $\|V - P_y Y\|_F^2$.

Given the data points of different modalities $X$, $Y$, and their latent representation $V$, we impose two additional constraints [13] which exploit the intra-media distribution of the data to preserve overall structure of the data.

- **Local Affinity**: If two data points are close in the original space, their embedded latent representations should also be close to each other.

- **Distant Repulsion**: If two data points are far apart in the original space, their embedded latent representations should also be far apart from each other.

We adopt the graph regularization framework [4] to preserve the local affinity constraint. Specifically, we minimize

$$\frac{1}{2} \sum_{i=1}^{N} \sum_{j=1}^{N} W_{ij}^a \|\mathbf{v}_i - \mathbf{v}_j\|^2 \quad (1)$$

where $W^a = [W_{ij}^a]$ is the affinity matrix. We define $W_{ij}^a = W_{ij}^{ax} + W_{ij}^{ay}$, where $W_{ij}^{ax} = \lambda_x \exp(-\|\mathbf{x}_i - \mathbf{x}_j\|^2)$ is the affinity between data points $\mathbf{x}_i$ and $\mathbf{x}_j$. Similarly, we define $W_{ij}^{ay} = \lambda_y \exp(-\|\mathbf{y}_i - \mathbf{y}_j\|^2)$ to be the affinity between data points $\mathbf{y}_i$ and $\mathbf{y}_j$. According to the inference in [4], eqn. (1) can be rewritten as $Tr(VL^aV^T)$, where $Tr(\cdot)$ denotes the trace of a matrix, and $L^a = D^a - W^a$ is the graph Laplacian. $D^a$ is a diagonal matrix whose entries are column sums of $W^a$ and $D_{ii}^a = \sum_i W_{ij}^a$.

We preserve the distant repulsion constraint as follows. Let $W_{ij}^r = W_{ij}^{rx} + W_{ij}^{ry}$, where $W_{ij}^{rx} = \lambda_x \|\mathbf{x}_i - \mathbf{x}_j\|^2$ is the distance between data points $\mathbf{x}_i$ and $\mathbf{x}_j$, and $W_{ij}^{ry} = \lambda_y \|\mathbf{y}_i - \mathbf{y}_j\|^2$ is the distance between data points $\mathbf{y}_i$ and $\mathbf{y}_j$. Here, we seek to minimize

$$\frac{1}{2} \sum_{i=1}^{N} \sum_{j=1}^{N} W_{ij}^r \exp(-\|\mathbf{v}_i - \mathbf{v}_j\|^2). \quad (2)$$

Eqns. (1) and (2) ensure data structure of each modality is retained in the projected space, and hence preserve the global structure of the data in the projected space. By jointly considering all constraints, projection matrices $P_x$ and $P_y$ can then be obtained by minimizing the objective function $\mathcal{L}$,

$$\begin{aligned}
\mathcal{L} &= \lambda_x \|X - U_x V\|_F^2 + \lambda_y \|Y - U_y V\|_F^2 \\
&+ \frac{\alpha}{2} \sum_{i,j} W_{ij}^a \|\mathbf{v}_i - \mathbf{v}_j\|^2 + \frac{\beta}{2} \sum_{i,j} W_{ij}^r \exp(-\|\mathbf{v}_i - \mathbf{v}_j\|^2) \\
&+ \mu \|V - P_x X\|_F^2 + \mu \|V - P_y Y\|_F^2 \\
&+ \gamma(\|V\|_F^2 + \|U_x\|_F^2 + \|U_y\|_F^2 + \|P_x\|_F^2 + \|P_y\|_F^2)
\end{aligned} \quad (3)$$

where $\gamma(\|V\|_F^2 + \|U_x\|_F^2 + \|U_y\|_F^2 + \|P_x\|_F^2 + \|P_y\|_F^2)$ are regularizers on the matrix variables to avoid overfitting.

### 3.3. Optimization

Objective function in eqn. (3) is non-convex with respect to the five matrix variables $P_x$, $P_y$, $V$, $U_x$, $U_y$. However, it is trivially observed that by fixing any four of the matrix variables as constants, eqn. (3) is convex with respect to the remaining variable. Towards this end, we learn $P_x$ and $P_y$ by optimizing eqn. (3) in an iterative framework, where in each iteration, we fix four variables as constants to learn a value for the remaining variable. Specifically, by fixing $P_y$, $V$, $U_x$, $U_y$ as constants and $P_x$ as variable, we can derive $\frac{\partial \mathcal{L}}{\partial P_x} = -2\mu V X^T + 2\mu P_x X X^T + 2\gamma P_x$. Consequently, by setting $\frac{\partial \mathcal{L}}{\partial P_x}$ to be zero, we can derive the update rule of $P_x$ as:

$$P_x = V X^T (X X^T + \frac{\gamma}{\mu} I)^{-1} \quad (4)$$

Similarly, by fixing $P_x$, $V$, $U_x$, $U_y$ as constants and $P_y$ as variable, we can obtain the update rule of $P_y$ as:

$$P_y = V Y^T (Y Y^T + \frac{\gamma}{\mu} I)^{-1} \quad (5)$$

The update rule for $U_x$ and $U_y$ can then be derived as

$$U_x = X V^T (V V^T + \frac{\gamma}{\lambda_x} I)^{-1}, \quad (6)$$

$$U_y = Y V^T (V V^T + \frac{\gamma}{\lambda_y} I)^{-1} \quad (7)$$

We define the affinity between $\mathbf{v}_i$ and $\mathbf{v}_j$ to be $W_{ij}^r \exp(-\|\mathbf{v}_i - \mathbf{v}_j\|^2)$. By fixing $P_x$, $P_y$, $U_x$ and $U_y$ to be constants and setting $\frac{\partial \mathcal{L}}{\partial V} = 0$, we can obtain:

$$\begin{aligned}
V(\alpha L + (2\mu + \gamma)I) + (\lambda_x U_x^T U_x + \lambda_y U_y^T U_y)V \\
= \lambda_x U_x^T X + \lambda_y U_y^T Y + \mu P_x X + \mu P_y Y
\end{aligned} \quad (8)$$

where $I$ is an identity matrix, $L = L^a - \frac{\beta}{\alpha} L^r$, and $L^r$ is the graph Laplacian of $W_{ij}^r \exp(-\|\mathbf{v}_i - \mathbf{v}_j\|^2)$ (computed using $V$ obtained in the previous step).

We show eqn. (8) has a unique solution as follows. We first simplify eqn. (8) to be a Sylvester equation $AV + VB = C$, where

$$A = \lambda_x U_x^T U_x + \lambda_y U_y^T U_y \quad (9)$$
$$B = \alpha L + (2\mu + \gamma)I \quad (10)$$
$$C = \lambda_x U_x^T X + \lambda_y U_y^T Y + \mu P_x X + \mu P_y Y \quad (11)$$

From [11], eqn. (8) has a unique solution if and only if the eigenvalues of $A$ and $B$ in eqns. (9) and (10) respectively satisfy $p_i + q_j \neq 0$ for all $i, j$, where $p_i$ and $q_j$ are eigenvalues of $A$ and $B$ respectively. Consider the definition of $A$ in eqn. (9). We observe that by setting parameters $\lambda_x$ and $\lambda_y$ of eqn. (3) to be greater than zero, $A$ is a sum of Gram matrix which is symmetric positive semi-definite. As such, $p_i \geq 0$ for all $i$. Similarly, we note that $L$ in eqn. (10) is a Laplacian matrix of a graph, and hence is also symmetric positive semi-definite. Given that $I$ is symmetric positive, with sufficient large value of $2\mu + \gamma$, $B$ in eqn. (10) is positive definite. As such, $q_j > 0$ for all $j$. Consequently, $p_i$ and $q_j$ satisfy $p_i + q_j \neq 0$ and thus eqn. (8) has a unique solution. Here, we solve eqn. (8) as $V = P\tilde{V}Q^{-1}$, where $\tilde{v}_{ij} = \frac{\tilde{c}_{ij}}{p_i + q_j}$, $\tilde{C} = P^{-1}CQ$, and $P$ and $Q$ are matrices such that $P^{-1}AP$ and $Q^{-1}BQ$ are diagonal matrices.

We summarize the proposed cross-media hashing method in Alg. 1. The main computation cost comes from updating $V$, which takes $O(N^3)$ time per iteration. In practice the proposed method typically converges within 50 to 100 iterations.

**Algorithm 1** Structure Preserving Collective Matrix Factorization Hashing

**INPUT:**
  Heterogeneous data: $X$ and $Y$
  Hash code length: $H$
**OUTPUT:**
  $P_x$ and $P_y$: projection matrices for modality $\mathcal{X}$ and $\mathcal{Y}$
1: Initialize $U_x, U_y, V, P_x, P_y$ by random matrices
2: **repeat**
3:   Update $V$ by Eqn. (8)
4:   Update $P_x$ by Eqn. (4)
5:   Update $P_y$ by Eqn. (5)
6:   Update $U_x$ by Eqn. (6)
7:   Update $U_y$ by Eqn. (7)
8: **until** convergence

### 3.4. Extensions

The above section details our cross-media hashing method for two modalities. It is straightforward to extend the method for heterogeneous data of multiple modalities as:

$$\mathcal{L} = \sum_g \lambda_g \|X^{(g)} - U_g V\|_F^2 + \sum_g \mu \|V - P_g X^{(g)}\|_F^2$$
$$+ \frac{\alpha}{2} \sum_{i,j} W_{ij}^a \|\mathbf{v}_i - \mathbf{v}_j\|^2 + \frac{\beta}{2} \sum_{i,j} W_{ij}^r \exp(-\|\mathbf{v}_i - \mathbf{v}_j\|^2)$$
$$+ \gamma(\|V\|_F^2 + \sum_g \|U_g\|_F^2 + \sum_g \|P_g\|_F^2)$$

where $X^{(g)}$ denotes the data matrix for modality $g$, $W_{ij}^a = \lambda_g \exp(-\|\mathbf{x}_i^{(g)} - \mathbf{x}_j^{(g)}\|^2)$, and $W_{ij}^r = \lambda_g \|\mathbf{x}_i^{(g)} - \mathbf{x}_j^{(g)}\|^2$. We also note that if the class labels of training data is available, eqn. (3) can be extended to be a supervised method by computing $W_{ij}^a$ and $W_{ij}^r$ with distance between class labels.

## 4. Experiments

We evaluate our technique on the challenging Wiki [16] and NUS-WIDE datasets [5] for two cross-media retrieval tasks: (a) image query versus text database, which seeks to retrieve semantically relevant text based on a query image, and (b) text query versus image database, which aims to retrieve semantically relevant image based on a query text. We compare our retrieval performance against the state-of-the-art multimodal hashing methods which obtained the best published results on the benchmark datasets. Specifically, we compared against unsupervised multi-modal hashing methods of Ding et al. [7] (termed as CMFH), Song et al. [17] (termed as IMH) and Gong et al. [8] (termed as CCA). CMFH utilizes collective matrix factorization to obtain unified latent representation for hash code generation, but does not preserve global data structure in their formulation, unlike our method. IMH generates intra-media consistency preserving hash codes with orthogonality constraints and has been evaluated against [12, 2] where it demonstrated superior performance. CCA maximizes the inter-media correlation using canonical correlation analysis. Unsupervised methods, such as IMH, CMFH, CCA and our proposed method, do not demand expensive class labelling of the training data, unlike supervised cross-media hashing methods. Correspondingly, given that multimedia data often do not have class label, unsupervised methods are more practical for real applications. Nevertheless, we compare against the recent supervised multimodal method of Gong et al. [8] (termed as CCA-3V) which has achieved notable success on the benchmark datasets. While the focus of this paper is cross-media retrieval, the method can be used in single-media retrieval and thus we also reported the performance for two single-media retrieval tasks: (c) image query versus image database, which seeks to retrieve semantically relevant image based on a query image, and (d) text query versus text database, which aims to retrieve semantically relevant text based on a query text.

## 4.1. Experimental settings

For all experiments in this paper, we use the following parameter settings. We compute the graph Laplacian with $k$ nearest neighbor, where $k$ is set as 5. We set the weight for local affinity as $\alpha = 100$, and that for distant repulsion as $\beta = 1$. The weights for matrix factorization of data modality $\mathcal{X}$ is set as $\lambda_x = 0.5$, and for data modality $\mathcal{Y}$ is as set as $\lambda_y = 0.5$. Finally, we define the projection matrices weight as $\mu = 100$ and the regularizer weight as $\gamma = 0.01$. These parameters are obtained through 3 fold cross-validation and we find the model performance to be only mildly sensitive to the parameter settings.

Reported results of other methods in this paper are obtained by the same parameter settings as those used in the methods. Additionally, given that methods CMFH [7] and proposed method are initialized with random matrices during training, we conducted 20 runs for each method, and report their average results in the paper. For all experiments, we adhere to the evaluation criteria of other methods, and, as closely as possible, use the same training and test instances as other methods for comparing performances. We report retrieval performance by the mean average precision (MAP) score. The MAP is computed by averaging the average precision (AP) values over all the queries in the query set. AP is calculated as $AP = \frac{\sum_{k=1}^{L} P(k) r(k)}{\sum_{k=1}^{L} r(k)}$, where $L$ is the total number of retrieved documents, $P(k)$ is the precision of the top-$k$ retrieved documents, and $r(k)$ is the relevance of the $k^{th}$ document. $r(k)$ is equal to 1 if the $k^{th}$ document is relevant to the query and equals to 0 otherwise. Large MAP values indicate better retrieval performance.

## 4.2. Retrieval Results on Wiki dataset

The Wiki dataset [16] is generated from Wikipedia's featured articles collection, and contains 2,866 image-text pairs. We used 2,000 image-text pairs for training. For each image-text pair, the text corresponds to an article describing an event or a person, and the image is semantically relevant to the content of the text. A text article is represented by a 10-dimensional topic distribution vector that is computed from latent Dirichlet allocation (LDA) model [1], and an image is represented by a 128-dimensional bag of words that is computed from SIFT descriptors [15]. These image and text features are the same as those used by the comparison methods. Each image-text pair is labelled with one of ten semantic classes (e.g. art, history). We do not use these labels during training, but instead use them only for evaluation: a test data is correctly retrieved if it has the same class label as the query data. We use 80% of the data as the database and the remaining 20% to form the query set.

Table 1 reports the MAP retrieval results of the proposed method with comparison to state-of-the-art methods across three hash code lengths of 16, 32 and 64. We report the corresponding precision-recall curves of the methods for cross-media retrieval tasks in Fig. 2. As observed, for all retrieval tasks, we obtain slightly better retrieval results than the state-of-the-art CMFH method [7] and outperform CCA, CCA-3V [8] and IMH [17] across all tested hash code lengths. We highlight that CCA and CCA-3V ignores the intra-media structure of the data to derive the multimodal hashing functions. While IMH and CMFH consider intra-media similarity constraint to derive the hashing functions, these methods ignore the dissimilarity between data points. Given that our proposed method exploits both local affinity and distant repulsion constraints (and couple these constraints with informative latent representations) and more effectively preserve the global structure of the data in the projected space, the performance difference is thus expected.

It is noteworthy that as the length of the hash codes increases, retrieval performance of our method improves while that of CCA, CCA-3V and IMH degrades. This is because CCA, CCA-3V and IMH impose orthogonality constraints on the hash codes and require each bit in the hash code to be independent of each other. Corresponding, hash bits in the initial bit positions exhibit largest possible variance (i.e. account for much of the variability of the data), whereas variance of hash bits from succeeding bit positions are monotonically decreasing [3]. In this aspect, as the code length increases, the hash codes will contain more bits with low variance which can be noisy [19], and thus lead to degraded performance. In contrast, our method which is based on matrix factorization leverages on latent matrices to embed the original data information. Given that longer hash codes derived from higher dimension latent matrices can encode greater extent of data information, it is thus not surprising that retrieval performance of our method improves with increasing length of the hash codes.

## 4.3. Retrieval Results on NUS-WIDE dataset

The NUS-WIDE dataset [5] comprises 269,648 images and their associated tags which are obtained from Flickr. Ground-truth class labels (e.g. sky, tree) for the image-tag pairs are also available in the dataset. Following the protocols of the comparison methods (CMFH, IMH), we pruned a subset of 186,577 image-tag pairs belonging to the 10 largest classes. Similar to Sect. 4.2, we used 2,000 image-text pairs for training, and do not exploit the class labels to train the models but for evaluation only. We represent each images by a 500-dimensional bag of words feature that is computed from SIFT descriptors, and its associated tag by a 1,000-dimensional tag occurrence feature vectors [5]. These image and text features are the same as those used by the other comparison methods. We divide this subset into two parts: 99% of the data is selected as the database set and the remaining 1% as the query set.

Table 1: MAP results on Wiki dataset [16]. "$\mathcal{X} \rightarrow \mathcal{Y}$" denotes "$\mathcal{X}$ query versus $\mathcal{Y}$ database retrieval task". The best results are shown in bold.

| Method | Image → Text Code Length | | | Text → Image Code Length | | | Image → Image Code Length | | | Text → Text Code Length | | |
|---|---|---|---|---|---|---|---|---|---|---|---|---|
| | 16 | 32 | 64 | 16 | 32 | 64 | 16 | 32 | 64 | 16 | 32 | 64 |
| CMFH [7] | 0.2299 | 0.2449 | 0.2537 | 0.2125 | 0.2286 | 0.2403 | 0.1442 | 0.1488 | 0.1524 | 0.5017 | 0.5266 | 0.5400 |
| IMH [17] | 0.1436 | 0.1369 | 0.1443 | 0.1302 | 0.1245 | 0.1286 | 0.1262 | 0.1251 | 0.1278 | 0.3221 | 0.3279 | 0.3604 |
| CCA [8] | 0.1930 | 0.1866 | 0.1773 | 0.1606 | 0.1425 | 0.1324 | 0.1277 | 0.1229 | 0.1207 | 0.3846 | 0.3479 | 0.3479 |
| CCA-3V [8] | 0.1794 | 0.1704 | 0.1684 | 0.1586 | 0.1476 | 0.1398 | 0.1287 | 0.1242 | 0.1214 | 0.4415 | 0.4253 | 0.4156 |
| Proposed method | **0.2432** | **0.2536** | **0.2598** | **0.2195** | **0.2345** | **0.2436** | **0.1482** | **0.1511** | **0.1534** | **0.5244** | **0.5453** | **0.5574** |

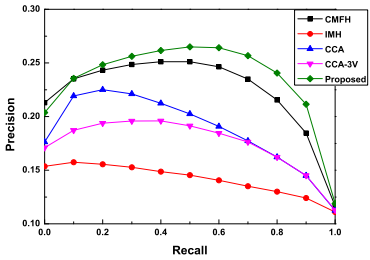

(a) Image Query v.s. Text Database @ 16 bits

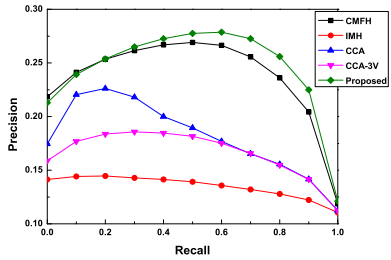

(b) Image Query v.s. Text Database @ 32 bits

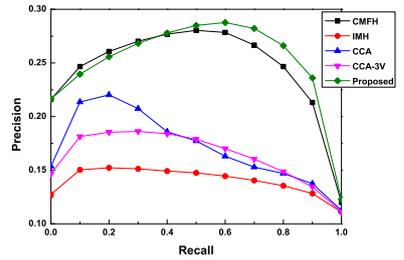

(c) Image Query v.s. Text Database @ 64 bits

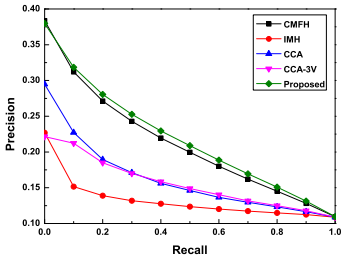

(d) Text Query v.s. Image Database @ 16 bits

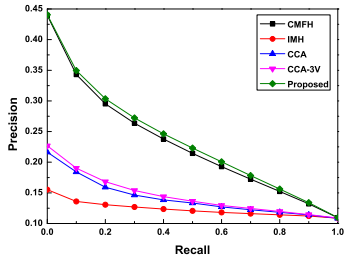

(e) Text Query v.s. Image Database @ 32 bits

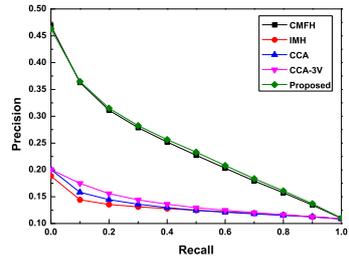

(f) Text Query v.s. Image Database @ 64 bits

Figure 2: Precision-recall curves for cross-media retrieval tasks on Wiki dataset (Best viewed on screen).

Table 2 reports the MAP retrieval results of the comparison and proposed methods across various hash code lengths. As observed, we obtained much higher MAP scores across all tested hash code lengths for both image-query and text-query retrieval tasks. The better performance of our method is also depicted by the precision-recall curves shown in Fig. 3, where we obtained higher precision scores at the evaluated recall values for cross-media retrieval tasks. We also reported the single-media retrieval performance in Table 2. As observed, we outperform existing methods for all tested code lengths. Importantly, we note that the setting in this experiment is similar in nature to real world scenario, in which the size of the training data is comparatively much smaller than the size of the test data. Correspondingly, retrieval performance of the proposed and comparison methods on this dataset is indicative of their actual performance in real scenarios.

## 5. Conclusions

In this paper, we propose a novel unsupervised hashing method for cross-media retrieval. We exploit local affinity and distant repulsion constraints to preserve global structure of the data. Specifically, we enforce the constraints that data points which are similar in the original feature space should be near each other in the projected space, while dissimilar data points should be apart from each other in the projected space. These constraints are incorporated with informative latent representation through matrix factorization to learn the hashing functions. As the objective function is

Table 2: MAP results on NUS-WIDE dataset [5]. "$\mathcal{X} \to \mathcal{Y}$" denotes "$\mathcal{X}$ query versus $\mathcal{Y}$ database retrieval task". The best results are shown in bold.

| Method | Image → Text Code Length | | | Text → Image Code Length | | | Image → Image Code Length | | | Text → Text Code Length | | |
|---|---|---|---|---|---|---|---|---|---|---|---|---|
| | 16 | 32 | 64 | 16 | 32 | 64 | 16 | 32 | 64 | 16 | 32 | 64 |
| CMFH [7] | 0.3668 | 0.3640 | 0.3629 | 0.3662 | 0.3636 | 0.3626 | 0.3618 | 0.3597 | 0.3589 | 0.3808 | 0.3788 | 0.3781 |
| IMH [17] | 0.3716 | 0.3685 | 0.3605 | 0.3706 | 0.3681 | 0.3600 | 0.3653 | 0.3678 | 0.3744 | 0.4197 | 0.4246 | 0.4177 |
| CCA [8] | 0.3559 | 0.3527 | 0.3495 | 0.3598 | 0.3552 | 0.3512 | 0.3686 | 0.3628 | 0.3576 | 0.3541 | 0.3524 | 0.3518 |
| CCA-3V [8] | 0.3570 | 0.3540 | 0.3512 | 0.3614 | 0.3574 | 0.3542 | 0.3679 | 0.3628 | 0.3590 | 0.3563 | 0.3537 | 0.3519 |
| Proposed method | **0.4005** | **0.4058** | **0.4093** | **0.3869** | **0.3911** | **0.3938** | **0.3755** | **0.3782** | **0.3799** | **0.4388** | **0.4469** | **0.4523** |

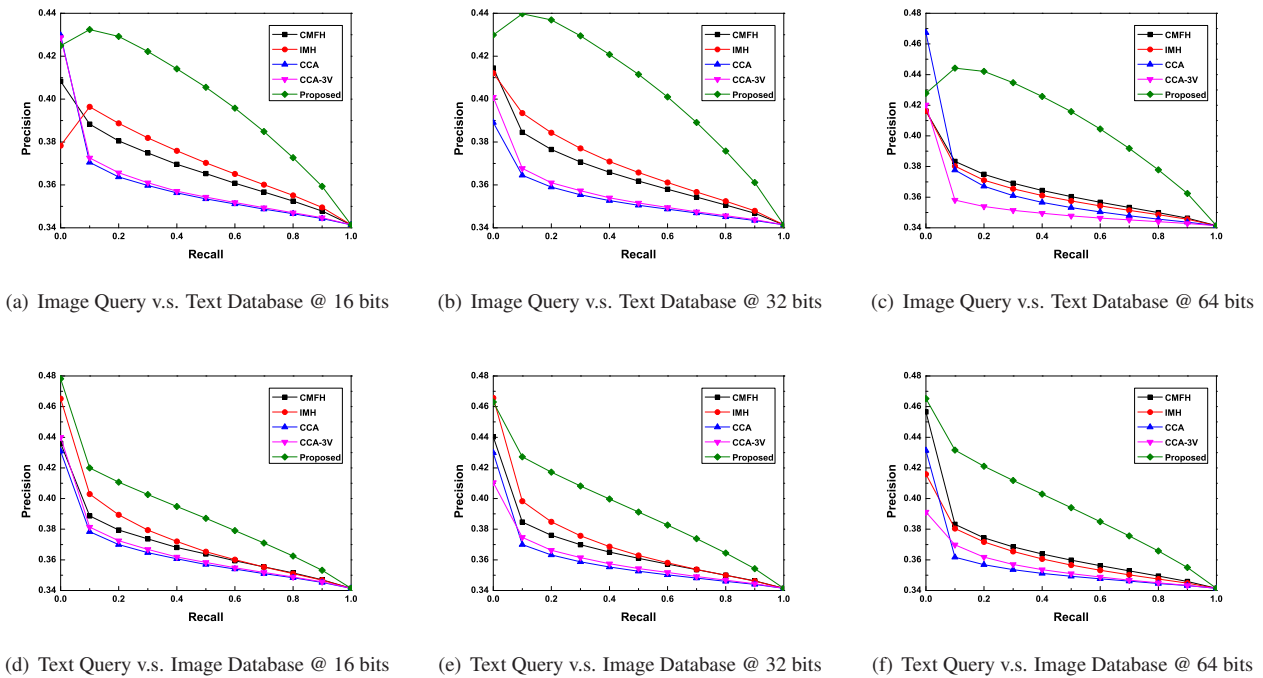

(a) Image Query v.s. Text Database @ 16 bits  
(b) Image Query v.s. Text Database @ 32 bits  
(c) Image Query v.s. Text Database @ 64 bits  
(d) Text Query v.s. Image Database @ 16 bits  
(e) Text Query v.s. Image Database @ 32 bits  
(f) Text Query v.s. Image Database @ 64 bits  

Figure 3: Precision-recall curves for cross-media retrieval tasks on NUS-WIDE dataset (Best viewed on screen).

non-convex, we learn the optimal hashing functions through an iterative framework. We evaluate our method on two popular multimedia datasets and compare against the state-of-the-art hashing methods. It is shown that the proposed method consistently outperforms the comparison methods on both single-media and cross-media retrieval tasks. As future work, we would like to devise more effective optimization techniques to improve the scalability of the proposed method. Additionally, we would also like to extend this work to multimodal image segmentation and alignment.

## References

[1] D. M. Blei, A. Y. Ng, and M. I. Jordan. Latent dirichlet allocation. *JMLR*, 3:993–1022, 2003. 6

[2] M. M. Bronstein, A. M. Bronstein, F. Michel, and N. Paragios. Data fusion through cross-modality metric learning using similarity-sensitive hashing. In *CVPR*, pages 3594–3601, 2010. 2, 3, 5

[3] C. J. Burges. *Dimension Reduction*. Now Publishers Inc, 2010. 6

[4] D. Cai, X. He, J. Han, and T. S. Huang. Graph regularized nonnegative matrix factorization for data representation. *TPAMI*, 33(8):1548–1560, 2011. 4

[5] T.-S. Chua, J. Tang, R. Hong, H. Li, Z. Luo, and Y. Zheng. Nus-wide: a real-world web image database from national university of singapore. In *ACM CIVR*, page 48, 2009. 5, 6, 8

[6] S. C. Deerwester, S. T. Dumais, T. K. Landauer, G. W. Furnas, and R. A. Harshman. Indexing by latent semantic analysis. *JASIS*, 41(6):391–407, 1990. 4

[7] G. Ding, Y. Guo, and J. Zhou. Collective matrix factorization hashing for multimodal data. In *CVPR*, pages 2083–2090, June 2014. 1, 2, 3, 5, 6, 7, 8

[8] Y. Gong, S. Lazebnik, A. Gordo, and F. Perronnin. Iterative quantization: A procrustean approach to learning binary codes for large-scale image retrieval. *TPAMI*, 35(12):2916–2929, 2013. 1, 2, 3, 5, 6, 7, 8

[9] J. He, J. Feng, X. Liu, T. Cheng, T.-H. Lin, H. Chung, and S.-F. Chang. Mobile product search with bag of hash bits and boundary reranking. In *CVPR*, pages 3005–3012, 2012. 1

[10] P. Indyk and R. Motwani. Approximate nearest neighbors: towards removing the curse of dimensionality. In *ACM STOC*, pages 604–613, 1998. 1

[11] A. Jameson. Solution of the equation ax+xb=c by inversion of an m*m or n*n matrix. *SIAM Journal on Applied Mathematics*, 16(5):1020–1023, 1968. 5

[12] S. Kumar and R. Udupa. Learning hash functions for cross-view similarity search. In *AAAI*, pages 1360–1365, 2011. 2, 5

[13] Z. Li, J. Liu, and H. Lu. Structure preserving nonnegative matrix factorization for dimensionality reduction. *CVIU*, 117(9):1175 – 1189, 2013. 4

[14] W. Liu, J. Wang, R. Ji, Y.-G. Jiang, and S.-F. Chang. Supervised hashing with kernels. In *CVPR*, pages 2074–2081, 2012. 2

[15] D. G. Lowe. Distinctive image features from scale-invariant keypoints. *IJCV*, 60(2):91–110, 2004. 6

[16] N. Rasiwasia, J. Costa Pereira, E. Coviello, G. Doyle, G. R. Lanckriet, R. Levy, and N. Vasconcelos. A new approach to cross-modal multimedia retrieval. In *ACM Multimedia*, pages 251–260, 2010. 5, 6, 7

[17] J. Song, Y. Yang, Y. Yang, Z. Huang, and H. T. Shen. Inter-media hashing for large-scale retrieval from heterogeneous data sources. In *SIGMOD*, pages 785–796, 2013. 1, 2, 3, 5, 6, 7, 8

[18] C. Wang and S. Mahadevan. Manifold alignment preserving global geometry. In *AAAI*, pages 1743–1749, 2013. 1

[19] J. Wang, S. Kumar, and S.-F. Chang. Sequential projection learning for hashing with compact codes. In *ICML*, pages 1127–1134, 2010. 2, 6

[20] Y. Weiss, A. Torralba, and R. Fergus. Spectral hashing. In *NIPS*, page 6, 2008. 2

[21] Z. Yu, Y. Zhang, S. Tang, Y. Yang, Q. Tian, and J. Luo. Cross-media hashing with kernel regression. In *ICME*, pages 1–6, July 2014. 3

[22] D. Zhang and W.-J. Li. Large-scale supervised multimodal hashing with semantic correlation maximization. *AAAI*, 2014. 1, 3

[23] Y. Zhen and D.-Y. Yeung. Co-regularized hashing for multimodal data. In *NIPS*, pages 1385–1393, 2012. 1, 2, 3